# Adaptive Optimization for Enhanced Efficiency in Large-Scale Language Model Training

Jiajing Chen
New York University
New York, USA

Bingying Liu
Independent Researcher
McLean, USA

Xiaoxuan Liao
New York University
New York, USA

Jia Gao
Stevens Institute of Technology
Hoboken, USA

Hongye Zheng
The Chinese University of Hong Kong
Hong Kong, China

Yue Li*
Purdue University
West Lafayette, USA

*Abstract*—With the rapid development of natural language processing technology, large-scale language models (LLM) have achieved remarkable results in a variety of tasks. However, how to effectively train these huge models and improve their performance and computational efficiency remains an important challenge. This paper proposes an improved method based on adaptive optimization algorithm, aiming to improve the training efficiency and final performance of LLM. Through comparative experiments on the SQuAD and GLUE data sets, the experimental results show that compared with traditional optimization algorithms (such as SGD, Momentum, AdaGrad, RMSProp and Adam), the adaptive optimization algorithm we proposed has better accuracy and F1 score. Both have achieved significant improvements, especially showed stronger training capabilities when processed large-scale texts and complex tasks. The research results verify the advantages of adaptive optimization algorithms in large-scale language model training and provide new ideas and directions for future optimization methods.

## I. INTRODUCTION

With the widespread application of large-scale language models (LLM) in the field of natural language processing (NLP), its performance in tasks such as text generation, sentiment analysis, and automatic translation has achieved remarkable results. However, although these models have demonstrated strong capabilities in multiple fields, LLM still faces some challenges in practical applications due to their large parameter scale and training complexity. Especially in scenarios with limited computing resources and high real-time requirements, how to improve the computational efficiency and inference performance of the model has become a hot issue in current research. Adaptive optimization algorithms [1], as a technology that can dynamically adjust learning strategies according to the characteristics of data and optimization goals in different tasks, have been proven to have significant advantages in many machine learning tasks [2-4]. Therefore, exploring the application of adaptive optimization algorithms in LLM is of great significance for improving model performance, reducing computing resource consumption [5], and achieving a more efficient inference process.

The advantage of the adaptive optimization algorithm is that it can dynamically adjust hyperparameters such as learning rate according to different situations encountered during model training, thereby making the training process more efficient and avoiding overfitting or training that may occur in traditional optimization algorithms [6]. In LLM training, due to the complexity of the model and the huge number of parameters, conventional optimization algorithms often require a large number of computing resources and lack flexibility in parameter adjustment, which often leads to problems such as long training time and unstable training effects [7]. The adaptive optimization algorithm can find a suitable learning path in a shorter time by dynamically adjusting the gradient change or loss function, thereby improving the training efficiency and performance of the model. Therefore, how to combine adaptive optimization algorithms with LLM and explore its effect on improving model performance has become an important research direction.

Combining adaptive optimization algorithms effectively improves LLM performance. Adaptive learning rates, momentum adjustments, and dynamic mechanisms stabilize training and accelerate convergence. These algorithms reduce overfitting by adjusting parameters based on model performance and optimize computing resources by intelligently adjusting calculation step sizes. Applying adaptive optimization enhances model performance, efficiency, and usability in real-world deployment [8].

LLM applications expand to intelligent customer service, automated content creation, and medical [9], financial [10], and other fields. However, with increased application scenarios, model real-time performance, and computational efficiency requirements rise. Traditional large-scale pre-training models consume significant computing resources during inference, posing challenges for high-real-time applications. Adaptive optimization algorithms enable LLM to achieve more efficient calculations, meeting real-time requirements. This research direction promotes LLM application in various fields and advances NLP model optimization and computational efficiency. Advanced adaptive optimization technology reduces computing resource dependence while ensuring model performance, accelerating the popularization of intelligent technology [11].

Research on LLM performance improvement based on adaptive optimization algorithms holds significant academic and practical value. It breaks through computing resource consumption and training efficiency bottlenecks, promoting wider and more popular application of large-scale language models. Improving optimization algorithms enhances LLM processing capabilities and adaptability to changing practical needs. This research is essential for technological progress and further development of intelligent and automated technology.

## II. RELATED WORK

Advancements in optimization techniques have significantly influenced the training of large-scale language models (LLMs), offering new avenues for enhancing computational efficiency and model performance. Numerous studies have explored dynamic learning strategies and their application in various machine learning contexts, providing a foundation for the adaptive optimization framework proposed in this work.

Tao et al. [12] introduced methodologies to enhance adaptability in large-scale models, demonstrating the utility of dynamic learning adjustments in improving classification accuracy and data synthesis capabilities. Similarly, Du et al. [13] leveraged advanced optimization methods in graph-based reasoning tasks, illustrating the potential for increased efficiency and stability in complex training environments.

Dynamic learning strategies have also been shown to improve interpretability and efficiency in high-dimensional data processing. Yan et al. [14] developed approaches to transform multidimensional data into interpretable representations, emphasizing the role of flexible learning rates and momentum adjustments. These findings align closely with the objectives of this study in improving the adaptability of optimization algorithms for LLM training.

Research on low-resource and few-shot learning further highlights the potential of adaptive methods to overcome challenges in data-scarce scenarios. Feng et al. [15] proposed integration strategies using generative models, demonstrating how adaptive mechanisms can enhance generalization capabilities and model robustness. Such techniques inform our exploration of scalable optimization algorithms for LLMs.

Multi-modal learning frameworks have similarly benefited from adaptive approaches. Duan et al. [16] and Liang et al. [17] explored multi-modal architectures, underscoring the importance of dynamic optimization in achieving stable learning across diverse data modalities. Metric learning, as discussed by Luo et al. [18], complements these efforts by addressing sparsity and improving adaptability in data representation tasks, further supporting the design of efficient optimization strategies for large-scale training. Additionally, Hu et al. [19] examined fine-tuning methods for domain-specific applications, highlighting how adaptive optimization can reduce computational overhead while preserving model performance, insights that are integral to the methodology presented in this paper.

Finally, the exploration of self-adaptive frameworks has provided valuable perspectives on optimizing convergence rates and stability in large-scale model training. Wang et al. [20] investigated automated adjustments in learning processes, showcasing the potential of adaptive strategies to enhance both efficiency and training outcomes. These contributions collectively inform the development of adaptive optimization algorithms tailored to the demands of large-scale language model training.

## III. METHOD

In this study, we proposed a scheme based on adaptive optimization algorithms to improve the performance of large-scale language models (LLMs) during training. In order to better adapt to different tasks and data characteristics, we combined adaptive optimization algorithms (such as Adam, AdaGrad, RMSProp, etc.) with traditional optimization methods to achieve more efficient training and inference processes. Specific methods include adaptive learning rate adjustment mechanism, optimization of gradient update rules [21], and adjustment of model structure. The algorithm framework diagram is shown in Figure 1

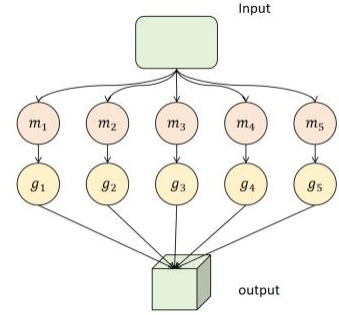

Figure 1 Algorithm framework diagram

First, we used the Adam (Adaptive Moment Estimation) algorithm in the adaptive optimization algorithm during the optimization process. The Adam algorithm combines the advantages of momentum and adaptive learning rate adjustment. It dynamically adjusts the learning rate of each parameter by introducing first-order moment estimation (i.e., the mean of the gradient) and second-order moment estimation (i.e., the variance of the gradient) [22], thereby avoiding the problems of unstable training and slow convergence caused by fixed learning rates in traditional optimization algorithms. The update rules of Adam are as follows:

$$m_t = \beta_1 m_{t-1} + (1-\beta_1) g_t$$

$$v_t = \beta_2 v_{t-1} + (1-\beta_2) g_t^2$$

$$m_t = \frac{m_t}{1-\beta_1^t}, \quad v_t = \frac{v_t}{1-\beta_2^t}$$

$$\theta_t = \theta_{t-1} - \eta \frac{m_t}{\sqrt{v_t} + \varepsilon}$$

Among them, $g_t$ is the current gradient, $m_t$ and $v_t$ are the first-order moment estimate and second-order moment estimate of the gradient respectively, $\beta_1$ and $\beta_2$ are the hyperparameters that control the attenuation of the first-order moment and the second-order moment respectively, $\eta$ is the learning rate, and $\varepsilon$ is a minimum value used to prevent zero division errors. In this way, the Adam algorithm can dynamically adjust the update step size of each parameter according to the gradient information during training, thereby improving the training efficiency.

In order to further improve the performance of LLM, we also introduced an adaptive learning rate adjustment mechanism. Usually, the learning rate at different stages of the training process should be different. The learning rate is larger in the early stage to accelerate convergence, and it should be gradually reduced in the later stage to avoid oscillation and overfitting. In this study, we used a learning rate decay strategy, specifically an exponential decay method, namely:

$$\eta_t = \eta_0 \cdot \gamma^t$$

Among them, $\eta_0$ is the initial learning rate, $\gamma$ is the decay rate, and t is the current number of training steps. Through this strategy, it can be ensured that in the later stage of training, the parameter update of the model becomes smoother, thereby helping the model to converge better and improve its generalization ability.

In addition, we have also improved the gradient update rule to deal with the gradient vanishing and gradient exploding problems that LLM often encounters during training. To overcome these problems, the gradient clipping technique is used. This method sets a threshold, and if the L2 norm of the gradient exceeds the threshold, the gradient is scaled to within the threshold. Specifically, if the gradient $g_t$ of a certain update satisfies:

$$// g_t //_2 > clip\_value$$

Then the updated gradient is:

$$g_t = g_t \cdot \frac{clip\_value}{// g_t //_2}$$

In this way, we can effectively prevent the occurrence of gradient explosion and ensure the stability of the training process.

In addition to the application of the above optimization algorithms, we also made appropriate adjustments to the structure of LLM to improve the training efficiency and inference performance of the model. Specifically, we adopt a multi-layer attention mechanism and introduce a sparsification strategy to reduce computational overhead. By dynamically selecting key features and ignoring redundant parts with enhanced feature extraction in heterogeneous information networks [23], we can reduce the computational effort while maintaining model performance without degradation. In addition, the use of Mixed Precision Training technology can significantly reduce memory usage and calculation time by converting some calculations into low-precision floating point numbers, such as LoRA-LiteE [24], further improving the training speed of the model.

During the experiment, we also adjusted the adaptive optimization algorithm according to the needs of different tasks. For example, in the text generation task, due to the large model output space, we added a regularization term to prevent the generated text from being too simple or overfitting the training data. In the question-and-answer task, we modified the loss function and introduced a weighting strategy based on the attention mechanism to make the model pay more attention to the key information in the question.

By combining adaptive optimization algorithms and model structure adjustments, this research aims to improve the performance of LLM so that it can perform tasks more efficiently in a variety of practical applications, especially when computing resources are limited, and still maintain a relatively high performance. High inference speed and accuracy.

IV. EXPERIMENT

A. Datasets

We selected the GLUE benchmark dataset, which contains nine subtasks evaluating natural language understanding capabilities, such as sentence pair classification, text entailment, and sentiment analysis. The GLUE dataset effectively evaluates the model's performance in various natural language understanding tasks and is highly representative.

Additionally, we selected the SQuAD dataset to further verify the model's performance in processing large-scale datasets. SQuAD contains a large amount of paragraph text and its corresponding questions and answers, requiring the BERT model to extract information from long articles to answer specific questions. Testing on the SQuAD dataset evaluated the performance improvement of our adaptive optimization algorithm on BERT in complex tasks, especially in terms of computational efficiency and accuracy when processing large amounts of text data.

Finally, an intuitive introduction table of the two datasets is given, as shown in Table 1.

Table 1 Experimental Results of SQuAD

| Dataset name | Task Type | Data size | Mission Statement |
|---|---|---|---|
| GLUE | Multitasking | More than 1M training samples | It includes sentence pair classification, text implication, sentiment analysis, etc., to evaluate various aspects of language comprehension. |
| SQuAD | Question and answer task | Over 100k question-answer pairs | Extract information from long paragraphs to answer questions and evaluate the model's comprehension and reasoning abilities. |

B. Experimental setup

We trained a BERT model using a standard training framework and reasonable hyperparameters. To simulate a large-scale language model's training, we selected key hyperparameters like batch size, learning rate, and optimization algorithm. We used a batch size of 32 for good model convergence speed and stability in natural language processing tasks. The initial learning rate was 2e-5, and Adam optimization algorithm was used for parameter updates. We combined adaptive learning rate decay with an exponential decay strategy to gradually decrease the learning rate during training, promoting smoother convergence in later stages.

In addition, considering the huge number of parameters in the BERT model, we limited the maximum number of steps (epochs) of training to 3 cycles. The training time for each cycle is about 2 hours. During the training process, the performance of the model on the validation set is regularly evaluated to ensure that the model does not overfit. During the training process, we also applied gradient clipping technology to prevent the gradient explosion phenomenon during gradient update and ensure the stability of training. To avoid overfitting, we also added a dropout layer to the model with a dropout rate of 0.1 to improve the generalization ability of the model.

In addition, to further improve the repeatability of the experiment and the stability of the results, we set the random seed of each experiment to 42 and used NVIDIA V100 GPU for accelerated training. Mixed precision calculation (FP16) was used during training to save video memory and speed up training. All experimental settings and codes are based on Hugging Face's Transformers library, ensuring efficient model loading, training, and evaluation processes. Through the adjustment and optimization of these hyperparameters, we can efficiently train the BERT model and evaluate the effect of different adaptive optimization algorithms on the performance of the model.

*C. Experimental result*

To validate the effectiveness of our adaptive optimization algorithm, we compared it with five traditional optimization algorithms: SGD, Momentum, AdaGrad, RMSProp, and Adam. [25] SGD, a widely used algorithm, updates parameters with a fixed learning rate but struggles with slow convergence and local optimality in large-scale models. Momentum accelerates gradient descent but requires hyperparameter adjustment. AdaGrad reduces learning rates dynamically but risks premature decay. RMSProp adjusts learning rates based on an exponential moving average but may lead to premature convergence. Adam combines Momentum and RMSProp's advantages, offering strong convergence and stability.

Our adaptive optimization algorithm is more flexible and efficient, dynamically adjusting learning strategies to avoid issues caused by fixed learning rates or parameter adjustments. It improves the training speed and final performance of the BERT model, especially in resource-constrained environments. Experiments were conducted on text classification tasks using the SQuAD and GLUE datasets. Results are presented in Tables 2 and 3.

Table 2 Experimental Results of SQuAD

| Model | Acc | F1 score |
|---|---|---|
| SGD | 75.2 | 77.1 |
| Momentum | 77.5 | 79.2 |
| AdaGrad | 76.3 | 78.0 |
| RMSProp | 78.0 | 79.5 |
| Adam | 80.1 | 81.2 |
| Ours | 82.4 | 83.1 |

Different optimization algorithms significantly affect the SQuAD dataset's model performance. Traditional SGD has low accuracy and F1 score, indicating slower training and a higher likelihood of local minima. Momentum and AdaGrad improve performance, balancing precision and recall. Momentum slightly outperforms AdaGrad, achieving 77.5% accuracy and 79.2% F1 score. RMSProp and Adam, adaptive optimization algorithms, enhance model stability and achieve 78.0% accuracy and 79.5% F1 score. Adam boosts performance to 80.1% and 81.2% F1 score. Adaptive optimization enhances model training efficiency and effectiveness. Our proposed adaptive optimization algorithm (Ours) achieves the largest improvement in accuracy (82.4%) and F1 score (83.1). Compared with Adam, ours has improved by 2.3% and 1.9% respectively in the two indicators, showing it can better dynamically adjust the learning strategy and adapt to the complexity and uncertainty in model training, especially in question-and-answer tasks with long texts and complex semantics. This result verifies the superiority of adaptive optimization in large-scale language model training and demonstrates its potential in practical applications.

Table 3 Experimental Results of GLUE

| Model | Acc | F1 score |
|---|---|---|
| SGD | 72.3 | 74.1 |
| Momentum | 74.2 | 75.8 |
| AdaGrad | 73.1 | 74.6 |
| RMSProp | 75.5 | 77.2 |
| Adam | 77.0 | 78.5 |
| Ours | 78.4 | 80.0 |

In the GLUE task, the performance of various optimization algorithms is slightly lower than that of SQuAD, indicating that the GLUE task is more difficult and covers multiple subtasks, so the model needs to show stronger adaptability and generalization ability in multiple fields. Compared with the SQuAD task, the GLUE dataset is more complex and diverse, involving tasks at multiple levels such as text reasoning, sentiment analysis, and sentence matching. Therefore, the subtle differences in the optimization algorithm have a more obvious impact on the final performance.

As can be seen from the table, the traditional SGD algorithm also performs relatively poorly in the GLUE task, with low accuracy and F1 scores of 72.3% and 74.1% respectively. The introduction of Momentum, AdaGrad and RMSProp has improved the model performance to a certain extent, especially the RMSProp algorithm, which has a relatively stable performance through the adjustment of the adaptive learning rate, with an accuracy of 75.5% and an F1 score of 77.2%. The Adam optimization algorithm showed a relatively strong advantage in the GLUE task, with an accuracy of 77.0% and an F1 score of 78.5%, which is much better than other traditional optimization methods.

However, the most outstanding results come from our proposed adaptive optimization algorithm (Ours), which achieved the most significant improvement in accuracy (78.4%) and F1 score (80.0), which is 1.4% and 1.5% higher than Adam respectively. This shows that our method can better adapt to the complexity of GLUE tasks, especially in multi-task learning environments, and can effectively improve the overall performance of the model, further verifying the advantages of adaptive optimization algorithms in large-scale language model training.

## V. CONCLUSION

In this study, we propose an improved method based on adaptive optimization algorithms and apply it to the training of large-scale language models (LLM). Through comparative experiments with traditional optimization algorithms (such as SGD, Momentum, AdaGrad, RMSProp, and Adam), the experimental results of our adaptive optimization algorithm on the SQuAD and GLUE data sets show significant advantages. Experimental results show that compared with other optimization methods, our proposed algorithm can better dynamically adjust the learning strategy and improve the training efficiency and final performance of the model. Especially in the SQuAD dataset and GLUE benchmark, our method achieves the highest improvement in accuracy and F1 score, verifying its potential in large-scale language model training.

Our algorithm performs well on multiple tasks, but there's room for optimization. First, adaptive optimization algorithms still rely on experience. Automating and optimizing hyperparameter adjustment is crucial. Our algorithm focuses on improving training and inference, but exploring multi-task learning and cross-task generalization is needed. Making the optimization algorithm more universal and adaptable to diverse natural language processing tasks will enhance the model's capabilities.

In the future, we plan to test the adaptive optimization algorithm in a wider range of application scenarios, especially in resource-constrained environments such as edge computing and large-scale language model applications on mobile devices. As computing power continues to improve, we can further explore more details in deep neural network training and improve the algorithm's performance in long text, complex reasoning, and real-time generation tasks. Ultimately, with the continuous improvement of adaptive optimization algorithms, we hope to promote the efficient deployment of large-scale language models in more practical applications and facilitate the development of intelligent technology.

## REFERENCES

[1] P. Sheilsspeigh, M. Larkspur, S. Carver, et al., "Dynamic Context Shaping: A New Approach to Adaptive Representation Learning in Large Language Models," 2024.
[2] J. Du, Y. Cang, T. Zhou, J. Hu, and W. He, "Deep Learning with HM-VGG: AI Strategies for Multi-modal Image Analysis", arXiv preprint, arXiv:2410.24046, 2024.
[3] Y. Wei, K. Xu, J. Yao, M. Sun, and Y. Sun, "Financial Risk Analysis Using Integrated Data and Transformer-Based Deep Learning", Journal of Computer Science and Software Applications, vol. 7, no. 4, pp. 1-8, 2024.
[4] Q. Sun, T. Zhang, S. Gao, L. Yang, and F. Shao, "Optimizing Gesture Recognition for Seamless UI Interaction Using Convolutional Neural Networks," arXiv preprint arXiv:2411.15598, 2024.
[5] X. Wang, X. Li, L. Wang, T. Ruan, and P. Li, "Adaptive Cache Management for Complex Storage Systems Using CNN-LSTM-Based Spatiotemporal Prediction", arXiv preprint, arXiv:2411.12161, 2024.
[6] Y. Zi, "Time-Series Load Prediction for Cloud Resource Allocation Using Recurrent Neural Networks", Journal of Computer Technology and Software, vol. 3, no. 7, 2024.
[7] Y. Xiao, "Self-Supervised Learning in Deep Networks: A Pathway to Robust Few-Shot Classification", arXiv preprint, arXiv:2411.12151, 2024.
[8] D. Kowieski, "Large-scale machine unlearning in transformer-based pre-trained language models: Evaluation and adaptation of existing machine unlearning frameworks KGA and SCRUB," unpublished.
[9] H. Liu, T. Zhou, Y. Xiang, A. Shen, J. Hu, and J. Du, "Enhancing Medical Image Segmentation with Deep Learning and Diffusion Models", arXiv preprint, arXiv:2411.14353, 2024.
[10] Z. Liu, X. Xia, H. Zhang and Z. Xie, "Analyze the Impact of the Epidemic on New York Taxis by Machine Learning Algorithms and Recommendations for Optimal Prediction Algorithms," Proceedings of the 2021 3rd International Conference on Robotics Systems and Automation Engineering, pp. 46-52, May 2021.
[11] Y. Yang, I. Li, N. Sang, L. Liu, X. Tang, and Q. Tian, "Research on Large Scene Adaptive Feature Extraction Based on Deep Learning", Preprints, doi: 10.20944/preprints202409.0841.v1, 2024.
[12] C. Tao, X. Fan, and Y. Yang, "Harnessing LLMs for API Interactions: A Framework for Classification and Synthetic Data Generation," arXiv preprint, arXiv:2409.11703, 2024.
[13] J. Du, G. Liu, J. Gao, X. Liao, J. Hu, and L. Wu, "Graph Neural Network-Based Entity Extraction and Relationship Reasoning in Complex Knowledge Graphs," arXiv preprint, arXiv:2411.15195, 2024.
[14] X. Yan, Y. Jiang, W. Liu, D. Yi, and J. Wei, "Transforming Multidimensional Time Series into Interpretable Event Sequences for Advanced Data Mining," arXiv preprint, arXiv:2409.14327, 2024.
[15] Y. Feng, A. Shen, J. Hu, Y. Liang, S. Wang, and J. Du, "Enhancing Few-Shot Learning with Integrated Data and GAN Model Approaches," arXiv preprint, arXiv:2411.16567, 2024.
[16] S. Duan, Z. Wang, S. Wang, M. Chen, and R. Zhang, "Emotion-Aware Interaction Design in Intelligent User Interface Using Multi-Modal Deep Learning," arXiv preprint, arXiv:2411.06326, 2024.
[17] A. Liang, "Enhancing Recommendation Systems with Multi-Modal Transformers in Cross-Domain Scenarios," Journal of Computer Technology and Software, vol. 3, no. 7, 2024.
[18] Y. Luo, R. Wang, Y. Liang, A. Liang, and W. Liu, "Metric Learning for Tag Recommendation: Tackling Data Sparsity and Cold Start Issues," arXiv preprint, arXiv:2411.06374, 2024.
[19] J. Hu, Y. Cang, G. Liu, M. Wang, W. He, and R. Bao, "Deep Learning for Medical Text Processing: BERT Model Fine-Tuning and Comparative Study," arXiv preprint, arXiv:2410.20792, 2024.
[20] S. Wang, Z. Liu, and B. Peng, "A Self-training Framework for Automated Medical Report Generation," Proceedings of the 2023 Conference on Empirical Methods in Natural Language Processing, pp. 16443-16449, December 2023.
[21] X. Chen, H. Zhou, and Y. Li, "Effective design space exploration of gradient nanostructured materials using active learning based surrogate models," Materials & Design, vol. 183, p. 108085, 2019.
[22] Z. Liu and J. Song, "Comparison of Tree-Based Feature Selection Algorithms on Biological Omics Dataset," Proceedings of the 5th International Conference on Advances in Artificial Intelligence, pp. 165-169, November 2021.
[23] J. Wei, Y. Liu, X. Huang, X. Zhang, W. Liu and X. Yan, "Self-Supervised Graph Neural Networks for Enhanced Feature Extraction in Heterogeneous Information Networks," 2024 5th International Conference on Machine Learning and Computer Application (ICMLCA), pp. 272-276, 2024.
[24] Y. Yang, C. Tao, and X. Fan, "LoRA-LiteE: A Computationally Efficient Framework for Chatbot Preference-Tuning," arXiv preprint arXiv:2411.09947, 2024.
[25] R. Zaheer and H. Shaziya, "A study of the optimization algorithms in deep learning", Proceedings of the 2019 Third International Conference on Inventive Systems and Control (ICISC), pp. 536-539, Jan. 2019.